\documentclass[letterpaper, 10 pt, conference]{ieeeconf}  

\IEEEoverridecommandlockouts                              

\usepackage{graphics} 
\usepackage{epsfig} 
\usepackage[linesnumbered,lined,ruled,commentsnumbered]{algorithm2e}
\usepackage{algpseudocode} 
\usepackage{cancel}
\usepackage{hyperref}
\usepackage[dvipsnames]{xcolor}
\usepackage{cite} 

\title{\LARGE \bf
Interaction of Autonomous and Manually-Controlled Vehicles: Implementation of a Road User Communication Service
}

\author{
    Nikita Smirnov, Sebastian Tschernuth, Walter Morales-Alvarez \emph{Student Member, IEEE},\\ and Cristina Olaverri-Monreal \emph{Senior Member, IEEE}
    \thanks{
        Johannes Kepler University Linz, Austria; Chair  Sustainable Transport Logistics 4.0. \texttt{\{nikita.smirnov, sebastian.tschernuth, walter.morales\_alvarez, cristina.olaverri-monreal\}@jku.at}
    }
}

\begin{document}

\maketitle{}
\thispagestyle{empty}
\pagestyle{empty}

\begin{abstract}

Communication between vehicles with varying degrees of automation is increasingly challenging as highly automated vehicles are unable to interpret the non-verbal signs of other road users. The lack of understanding on roads leads to lower trust in automated vehicles and impairs traffic safety. To address these problems,  we propose the Road User Communication Service, a software as a service platform, which provides information exchange and cloud computing services for vehicles with varying degrees of automation. 
To inspect the operability of the proposed solution, field tests were carried out on a test track, where the autonomous JKU-ITS research vehicle requested the state of a driver in a manually-controlled vehicle through the implemented service. The test results validated the approach showing its feasibility to be used as a communication platform. A link to the source code is available.

\end{abstract}

\section{INTRODUCTION}



As the level of automation for new vehicles increases, the interaction between vehicles of lower and higher levels of autonomy could become more difficult, because driverless vehicles are not capable of acknowledging eye contact as a non-verbal sign from other drivers in the vicinity. This has been shown to affect the confidence in self-driving cars ~\cite{Olaverri-Monreal2020} and it is still unclear how the interaction between fully autonomous vehicles and other road users will be~\cite{Habibovic2019} \cite{Morales-Alvarez2020_2}.
In this context, other ways of communicating might increase the chances of gaining cooperation.


Connected vehicles could leverage inter-vehicular communication to enlarge the data acquired from the sensors with data received from the network for a more robust and complete perception of the environment. In this line of research, the purpose of the internet of vehicles (IoV) is to enable seamless communication between road users. The IoV utilizes Vehicle to Everything (V2X) as the technology enabling wireless communication on roads. It includes smart road infrastructure, connected vehicles, technologies for sharing data, and real-time data computation. V2X communication, in turn, provides connected vehicles with the necessary tools to interact with the road environment, namely road infrastructure (V2I), pedestrians (V2P), and other vehicles (V2V). V2X communication also includes a vehicle-to-network (V2N) connection. Currently, the most common V2X technologies are dedicated short-range communication (DSRC) and cellular V2X (C-V2X) \cite{Habibovic2019}. As a crucial component of the IoV, car cloud computing can augment communication capabilities through a low-latency information transfer between vehicles.
The communication technologies described above set the foundation for the method implemented in this paper to facilitate the coexistence of manually-controlled and highly automated vehicles. 


We contribute to the body of research in the field of automated vehicles and propose an approach that, to the best of our knowledge has not been explored previously in the literature. We implement a Road User Communication Service (RUCS) as a software as a service (SaaS) platform designed to enable the communication between vehicles with any levels of automation and also other road users that are equipped with a smart device. We validate our framework through a field test in which an autonomous vehicle utilizes RUCS for requesting information about a vehicle in the adjacent lane before making the pertinent decision regarding a lane-change manoeuver. The proposed platform attempts to make non-verbal communication unambiguous and interpretable to all road users.

The remainder of this paper is organized as follows. Section \ref{sec:RelatedWork} describes related work in the field of communication and human-machine interaction. In Section \ref{sec:RUCS}, we describe the RUCS platform in detail. Section \ref{sec:SystemEvaluation} describes the implemented field test scenario and the data collection procedure. Section \ref{sec:Results} presents the results of the  experiment. The last section discusses limitations and potential improvements of the platform, concludes the paper, and outlines future steps.

\section{Related Work}
\label{sec:RelatedWork}

In recent literature, several technologies have been presented to enhance driving perception and help increase road safety through approaches that augment the drivers' awareness of the surrounding environment. For example, the DSRC-based See-Through System~\cite{Olaverri-Monreal2010} enhanced the driver's visibility in case of a leading vision-obstructing vehicle, thus supporting the drivers overtaking decision.
Efficient communication has the potential to promote trust in AVs as stated in~\cite{Olaverri-Monreal2020}. The effects of external vehicle interfaces for autonomous vehicles (AVs) on the perceived safety, trust and acceptance of other road users have been mostly investigated by focusing on pedestrians (e.g.~\cite{9310642, alvarez2020autonomous, de2019perceived, burns2019pedestrian}) and only a few works studied interaction with conventional vehicles. In this context, for example, the results of a series of workshops and interviews involving truck drivers with platooning experience as well as passenger vehicle drivers, original equipment manufacturers (OEMs) and other experts in the field were presented in~\cite{Habibovic2019}. The results indicated that the most relevant situations for external signaling for platoons were highways at on-ramps, off-ramps, during overtaking, and lane changes. The same study refers to the work in~\cite{kitazaki2015effects} that states that traffic safety and efficiency, as well as trust in AVs may be enhanced if AVs are able to communicate information on their state and information.
To this end, a potential communication paradigm could be the one presented in~\cite{olaverri2017effect, olaverri2016tailigator}, a display-based system that was used to convey information regarding a potentially unsafe road situation between two following vehicles independently of their communication capabilities or equipment. In this context, various driving behavioral patterns from the following vehicle were observed depending on whether the system was located in the following vehicle or the rear part of the leading vehicle.\\
Regarding the development of V2X communication, the authors of \cite{Zhou2020} reviewed several communication protocols, including the 802.11p IEEE standard. They also investigated the development of cloud-based IoV (C-IoV) applications, which could, for instance, enable big data analysis, and described several architectures.\\
Further, a vehicular ad-hoc network (VANET)-based cloud model, which sought to integrate cloud computing into VANETs, was discussed in \cite{Bitam2015} at a theoretical level. 
Interoperability was regarded to be one key challenge. \\
To address data privacy, the study presented in \cite{Garg2019} proposed a smart security framework for VANETs, using edge computing nodes and the global wireless standard 5th generation mobile network 5G. The implemented solution demonstrated that latency, privacy, and security issues of cloud computing in VANETs can be improved by incorporating edge nodes.\\
In \cite{Lee2014} a further study described a vehicular cloud networking (VCN) system combining information-centric networking (ICN) and vehicular cloud computing (VCC). VCC allows users to form a vehicle-cloud, on which they can share information. ICN is responsible for the efficient distribution of the data among the users.\\
A model that establishes secure and efficient computing resource allocation between heterogeneous vehicles and road side units (RSUs) was presented in \cite{Lin2018}. Based on semi-Markov decision processes, this model considered the heterogeneity of vehicles and RSUs.\\
Although all these studies present relevant results for modeling communication networks or transferring information between road users, they either focus only on pedestrians, or lack cross-validation in real environments. Therefore, the main contribution of this work encompasses the design and implementation of a SaaS platform to enable the communication between vehicles with any level of automation and other road users equipped with a smart device. We additionally validate our RUCS framework through a field test as described in Section~\ref{sec:SystemEvaluation}.



\section{Description of the implemented Road User Communication Service}
\label{sec:RUCS}

\begin{figure}[ht]
	\centering
	\includegraphics[width=0.3\textwidth]{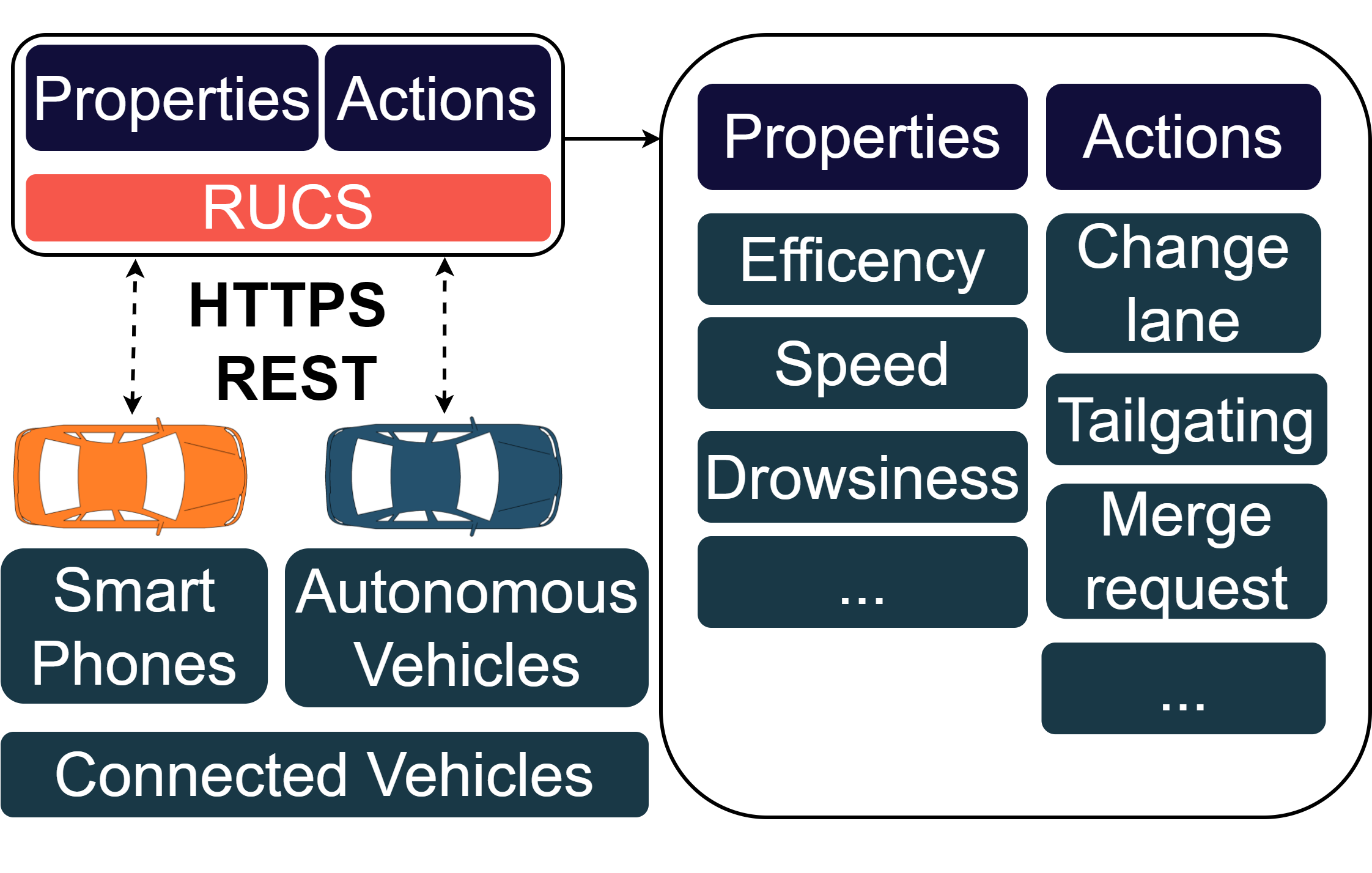}
	\caption{A simplified illustration of the interaction between RUCS and clients} 
	\label{fig:rucs_main}
\end{figure}

The main target of the implemented RUCS platform is to enable interactions between autonomous vehicles and other road users, including vulnerable road users and manually-driven vehicles. For that purpose, the service provides a reliable way of information exchange between connected vehicles and smart devices via representational state transfer (REST) application programming interface (API), Hypertext Transfer Protocol Secure (HTTPS) and Advanced Message Queuing Protocol (AMQP) \cite{AMQPWorkingGroup2008}, which are well-established protocols in high-load projects.

 It additionally provides tools to compute resource-intensive tasks in the cloud. Figure 1 illustrates the interaction process between the service and the clients. 

To separate the service from the client application, which may be written in a different language and run on various platforms, RUCS relies on the REST API protocol, which is independent of the specific client application. Since the application layer uses the service to transfer the data, the RUCS does not depend on the communication layer, therefore working with any V2N technology.

The service contains two types of interactions: 

\begin{itemize}
\item Properties: They can be requested via the service by the client. When a property is requested from a nearby vehicle, the service returns the correspondent value to the client. A property, therefore, is data computed based on a specific road user's state.
\item Actions: They include forwarding a message from the client to a vehicle that made a request so that it can react accordingly. In other words, requesting an action is an explicit way of asking for something from another road user.  
\end{itemize}

The following subsections describe the stack of technologies, the basic requests, the relationships between tables and the server-side algorithms of properties and actions.

\subsection{Stack of technologies} 

\begin{figure}[!t]
	\centering
	\includegraphics[width=0.3\textwidth]{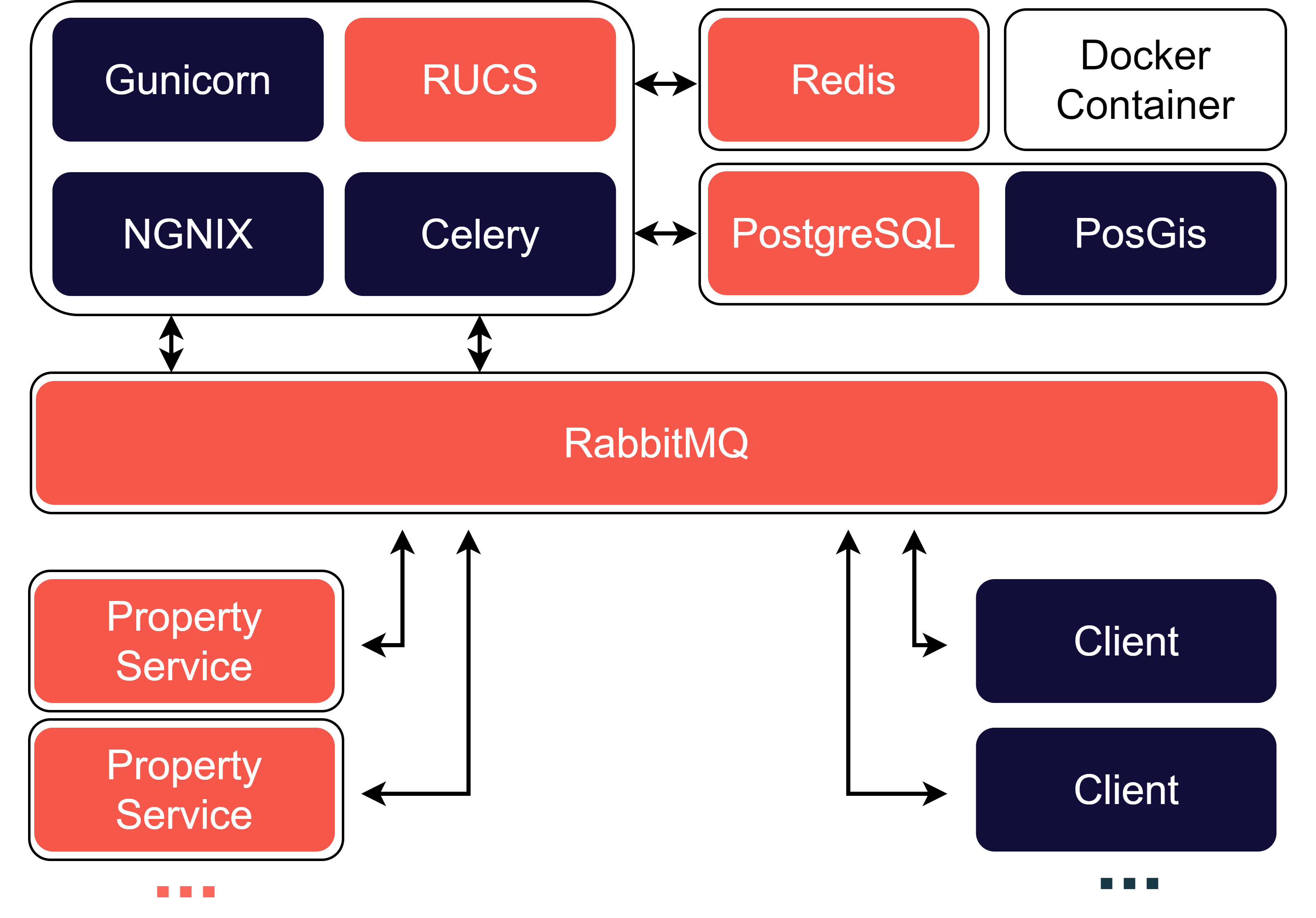}
	\caption{Architecture of the proposed service} 
	\label{fig:rucs_tech}
\end{figure}

In Figure~\ref{fig:rucs_tech}, we present a simplified scheme of the technologies on which we relied to implement the proposed service. The service worked on Python 3, Django framework \cite{Vincent2019a}, and the Django-rest-framework \cite{Vincent2019} extension to significantly simplify the prototype development. PostgreSQL \cite{Riggs2019} was selected as the database management system, as it is a highly reliable solution that is well-integrated within Django. To work with distributed data, we installed the PostGIS \cite{Obe2011} extension on top of PostgreSQL. We also used the key-value database Redis \cite{Carlson2013} as a cache to reduce the load from the PostgreSQL database. We utilized RabbitMQ \cite{Videla} as a message broker to transfer messages between clients. RabbitMQ adopts the AMQP protocol, for message transmission of high-load applications. Afterwards, we implemented the Celery \cite{Solem2013} technology to compute resource-intensive properties on third-party services. Celery relies on RabbitMQ and Redis technologies. We then relied on Docker \cite{Merkel2014} as a tool for containerization and virtualization to deploy the service prototype. Finally, we selected Gunicorn \cite{Carlson2013} as a web server gateway interface (WSGI) \cite{Gardner2009} and NGNIX \cite{Dejonghe2020} as a web service.

\subsection{General data flow}
\label{sec:Genera_data_flow}

\begin{figure}[!t]
	\centering
	\includegraphics[width=0.3\textwidth]{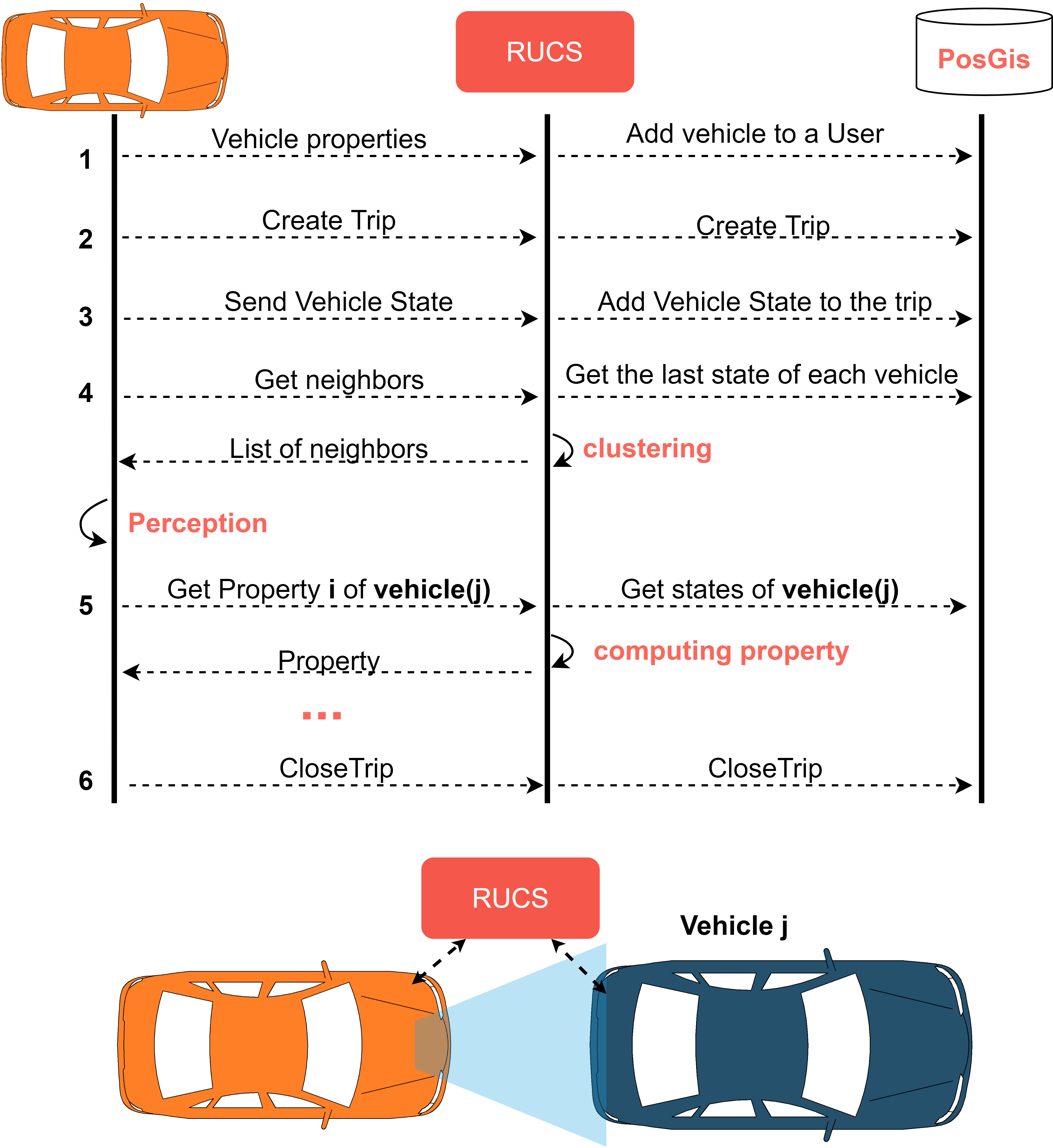}
	\caption{UML sequence diagram of a RUCS service and a client vehicle} 
	\label{fig:rucs_seq}
\end{figure}

Figure~\ref{fig:rucs_seq} shows the process for the data transmission after a request has been performed. The request types are detailed as follows:

\textbf{User registration}. It requires information about users' credentials and vehicles. Users complete a web form, in which they specify their vehicle model, year, plate number, color, etc. To enhance data protection, they also select properties and actions, which other users can request from the neighboring vehicles. After registration, the users get access to the functionality of the RUCS.

\textbf{Trip activation}. The service operates with the trip id instead of the user's id to increase data privacy. The client application sends a trip creation request when a user starts to drive. The service returns two topics in response to the trip creation request. One topic is for listening to messages from the service and another one for sending events to it.

\textbf{Vehicle states transferring}. GPS data of the vehicle's user as well as other vehicle states can also be transmitted. This information includes the state of the control system, the engine, and the driver's state (e.g. drowsiness) within a fixed interval. The service handles and logs the data for subsequent properties calculation.

\textbf{Neighbors request}. A client application requests a list of neighbors within a specified radius with a fixed interval. For each neighbor, the list includes information about their trip ID, vehicle description, GPS data, and a set of properties and actions, which the user can request from a neighbor.

\textbf{Property or action request}. The user can request a property or an action of a specific neighbor. For that purpose, the user sends a request with the neighbor's trip id and property or action of interest. If a property is requested, the service computes its content based on previously stored states from vehicles in the vicinity that are part of the request. In the case of an action request, the service will send the requested command to the listening topic of a requested neighbor. The neighbor's answer is then forwarded back to the user.

\textbf{Trip completion}. After having arrived at the destination, the user sends a request to close the trip. \\
It is noteworthy that connected vehicles can eliminate the need for the driver to interact with the service. A connected automobile can take full responsibility for vehicle-service communication and provide the driver with only high-level information.

\subsection{Database structure}

\begin{figure}[!t]
	\centering
	\includegraphics[width=0.3\textwidth]{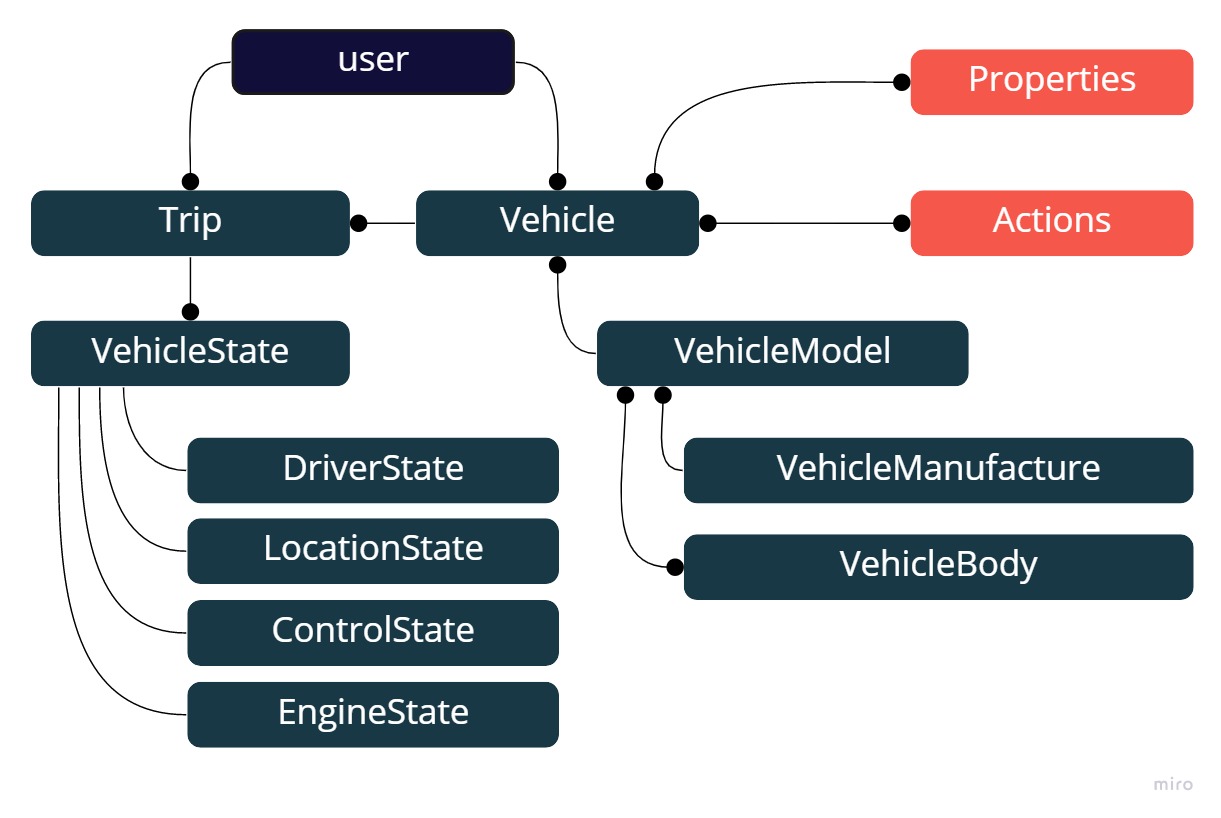}
	\caption{UML data model diagram of RUCS service} 
	\label{fig:rucs_db}
\end{figure}

Figure~\ref{fig:rucs_db} illustrates the RUCS data structure through a data model diagram. The components are arranged as follows. A user can have several vehicles. Each of them has a model, year of manufacture, plate number, and a set of properties and actions that the user selects from the list provided by the service. The initially-selected properties and actions can be requested by other users and will appear in the neighbors request (see \ref{sec:Genera_data_flow}). The user also has an unlimited number of trips, which they create requesting a trip activation. All the states the service receives from users are associated with their activated trips, thus ensuring the data privacy of each user. It is mandatory to provide the LocationState when updating the VehicleState due to the fact that the RUCS service uses these data for computing neighbors near the user's vehicle. Other states, such as the ControlState, are used only for property computation. Therefore providing these states is optional, and they are only transmitted if the client application has access to the vehicle data.

\subsection{Property and Actions}

\begin{figure}[!t]
	\centering
	\includegraphics[width=0.3\textwidth]{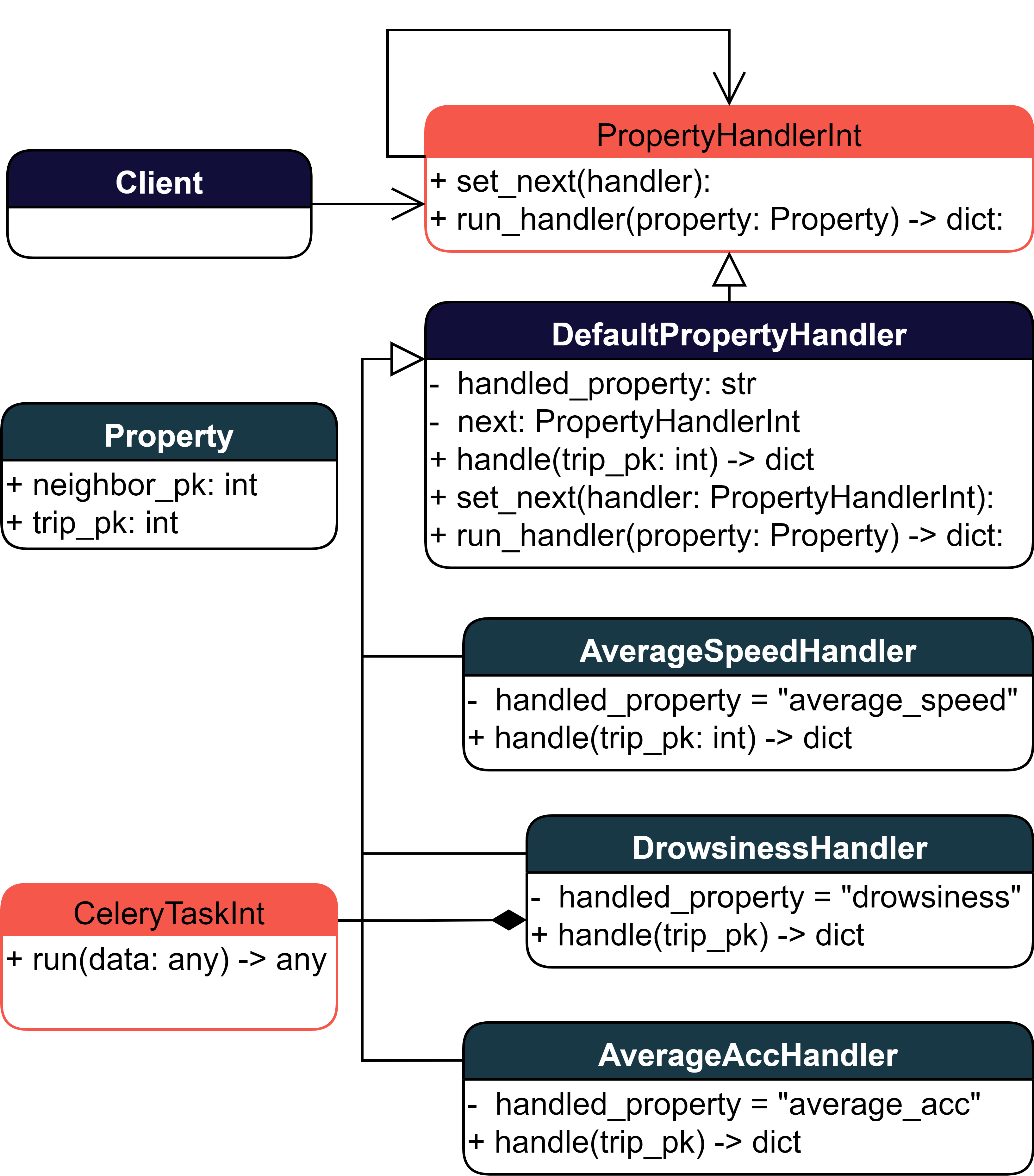}
	\caption{UML class diagram of the property handler} 
	\label{fig:rucs_handler}
\end{figure}

\begin{algorithm}
\label{code:property_handler}
    
    \caption{Property Request Handler}
    \small
    \KwIn{**kwargs \# request properties}    
    \KwOut{response in a JSON format}
    \SetKwData{NeighbourProperty}{property}
    \SetKwData{Handler}{handler}
    \SetKwData{PropertyDict}{p\_value}
    \SetKwData{Serializer}{serializer}
    \SetKwData{SerializerError}{serializer.error}
    \SetKwData{SerializerData}{serializer.data}
    
    \SetKwFunction{Property}{Property}
    \SetKwFunction{GetHandler}{get\_handler}
    \SetKwFunction{RunHandler}{handler.run\_handler}
    \SetKwFunction{GetSerializer}{get\_serializer}
    \SetKwFunction{Response}{Response}
    \SetKwFunction{SerializerIsValid}{serializer.is\_valid}
    
    \BlankLine
    
    \NeighbourProperty$\leftarrow$ \Property{$**kwargs$}\;
    \Handler$\leftarrow$ \GetHandler{}\;
    \PropertyDict$\leftarrow$ \RunHandler{\NeighbourProperty}\;
    \Serializer$\leftarrow$ \GetSerializer{}\;
    
    \BlankLine
    \If{\textbf{not} \SerializerIsValid{}}{
        \textbf{return} \Response{\SerializerError}\;
    }
    
    \BlankLine
    \textbf{return} \Response{\SerializerData}\;

\end{algorithm}

\begin{algorithm}
\label{code:action_handler}
    
    \caption{Action Request Handler}
    \small
    \KwIn{request, **kwargs \# request properties}    
    \KwOut{response status code}
    \SetKwData{NeighbourProperty}{property}
    \SetKwData{Topic}{topic}
    \SetKwData{UserKey}{user\_key}
    \SetKwData{Action}{action}
    
    \SetKwFunction{Property}{Property}
    \SetKwFunction{Exists}{cache.exists}
    \SetKwFunction{CacheGet}{cache.get\_topic}
    \SetKwFunction{TripGet}{Trip.get\_topic}
    \SetKwFunction{GetAction}{get\_action}
    \SetKwFunction{RabbitSend}{rabbit.send}
    \SetKwFunction{GetUserKey}{get\_user\_key}
    
    \BlankLine
    
    \NeighbourProperty$\leftarrow$ \Property{$**kwargs$}\;
    \UserKey$\leftarrow$ \GetUserKey{$request$}\;
    
    \BlankLine
    
    \eIf{\Exists{\UserKey}}{
        \Topic$\leftarrow$ \CacheGet{\UserKey}\;
    }{
        \Topic$\leftarrow$ \TripGet{\UserKey}\;
    }
    \BlankLine
    
    \If{\textbf{not} \Topic}{
        \textbf{return} $400$\;
    }
    
    \BlankLine
    
    \Action$\leftarrow$ \GetAction{\NeighbourProperty}
    \RabbitSend{\Topic, \Action}
    
    \BlankLine
    
    \textbf{return} $200$\;

\end{algorithm}

To process a property request, we used the chain-of-responsibility design pattern as described in Figure~\ref{fig:rucs_handler}, which allows the usage of several property handlers. Furthermore, the handler can use composition with Celery task classes to calculate resource-intensive tasks on third-party services. This architecture makes it possible to balance the service load and encapsulate the logic of handling properties from a client as Algorithm~\ref{code:property_handler} shows. It also defines an interface for third-party companies to design particular properties and integrate them into our service.
The service uses RabbitMQ to forward actions (see the action request handler in Algorithm~\ref{code:action_handler}). All action requests pass through the service to hide the logic implementation and the process validation. When the service receives an action request, it retrieves a listening topic of the requested vehicle from Redis and sends the action into the retrieved topic. The requested vehicle receives the action, executes it, and then responds, sending an action response back to the requesting user via the service. This architecture adds extra delays in the transmission of messages. However, it allows us to encapsulate the logic of the transactions on the service side and have a complete understanding of the communications on the road.


\section{System Evaluation}
\label{sec:SystemEvaluation}

\begin{figure}[!t]
	\centering
	\includegraphics[width=0.3\textwidth]{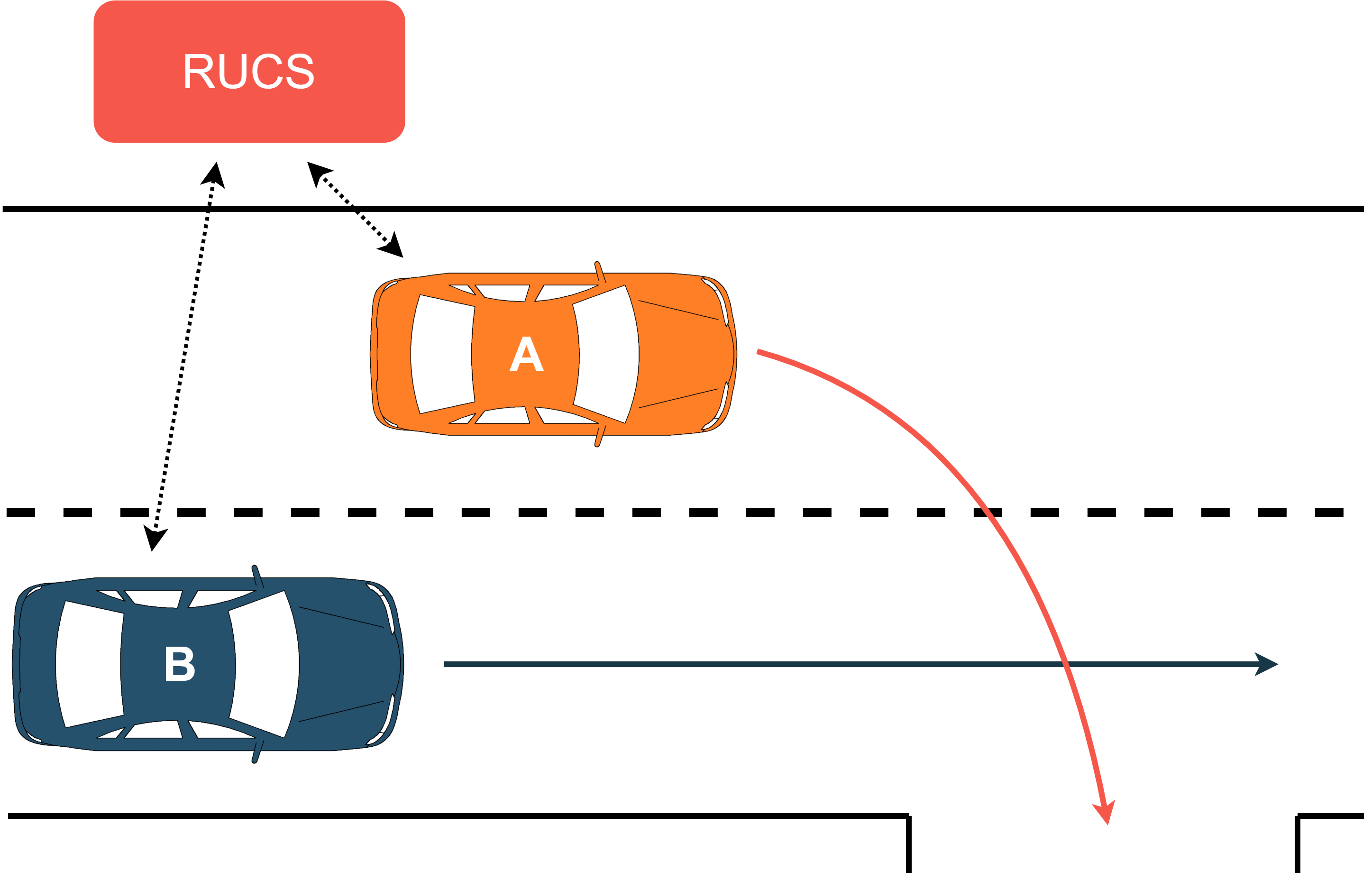}
	\caption{Illustration of the scenario to conduct the field text. A and B represent the autonomous and manually-driven vehicle respectively.} 
	\label{fig:scenario}
\end{figure}

The main goal of the RUCS platform is to enable communication between road users with different levels of automation. We evaluated the system to find out potential delays between requests and responses.

\subsection{Field test scenario and conditions}

\begin{figure}[!t]
	\centering
	\includegraphics[width=0.4\textwidth]{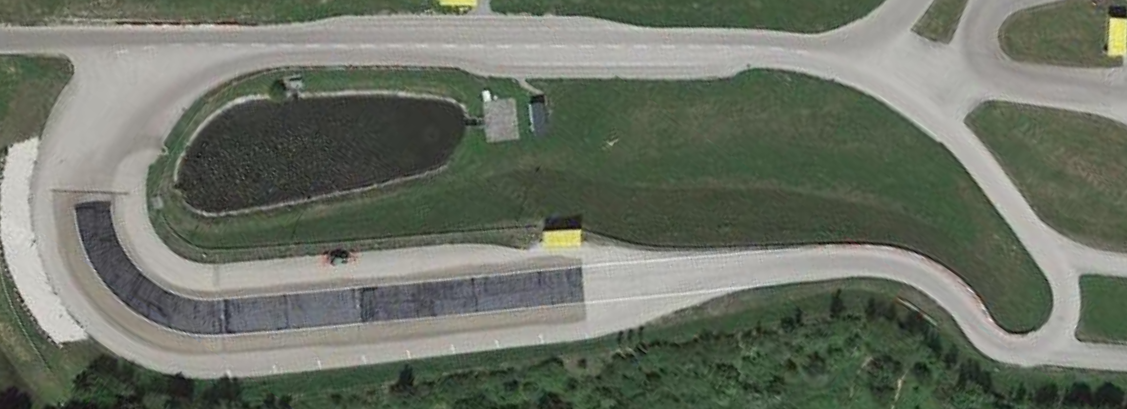}
	\caption{Satellite view on the test track in  \"OAMTC Marchtrenk} 
	\label{fig:track}
\end{figure}

To validate the proposed system we implemented a use case in which a low-cost, smartphone-based Driver State Monitoring System (DSMS) located in the manual-controlled vehicle had gathered information regarding drowsiness~\cite{allamehzadeh2017cost} and sent it to the cloud. 
We tested the message transfer by conducting several field tests in the \"OAMTC Driving Technology Centre in Marchtrenk, Upper Austria. Figure~\ref{fig:track} shows the test track. Realistic urban traffic conditions were achieved by driving at approximately 50 km/h. 
The source code of the RUCS platform is available.
\footnote{\url{https://gitlab.com/intelligent-transportation-systems/iamcv/communication_service}}

The autonomous vehicle A drove in the left lane. Vehicle B drove in the right lane and behind vehicle A (Figure~\ref{fig:scenario}). Vehicle A had to turn right to exit the road and needed therefore to perform a lane change. The driver in vehicle B was controlled by a driver that presented signs of drowsy driving. The autonomy of vehicle A was guaranteed by using the research vehicle from the Chair for Sustainable Transport Logistics 4.0 at the Johannes Kepler University in Linz (JKU-ITS)~\cite{Certad2022}. Figure~\ref{fig:jku-its-vehicle} shows the two vehicles during the field test. The autonomous vehicle A (monitored constantly by a team member) used the communication service to request the state of the driver (drowsy/non-drowsy) in vehicle B and perform the lane change accordingly. If the driver's state in vehicle B was drowsy, vehicle A would decelerate or wait until vehicle B was in front to then change the lane behind it. For the experiments, the service prototype was deployed in the amazon web services cloud using the Amazon Elastic Compute Cloud product for hosting and the ``t3a.large'' virtual machine. The data center was located in Frankfurt as it was the nearest location to the test track.

\subsection{Client description}

\begin{figure}[!t]
	\centering
	\includegraphics[width=0.24\textwidth]{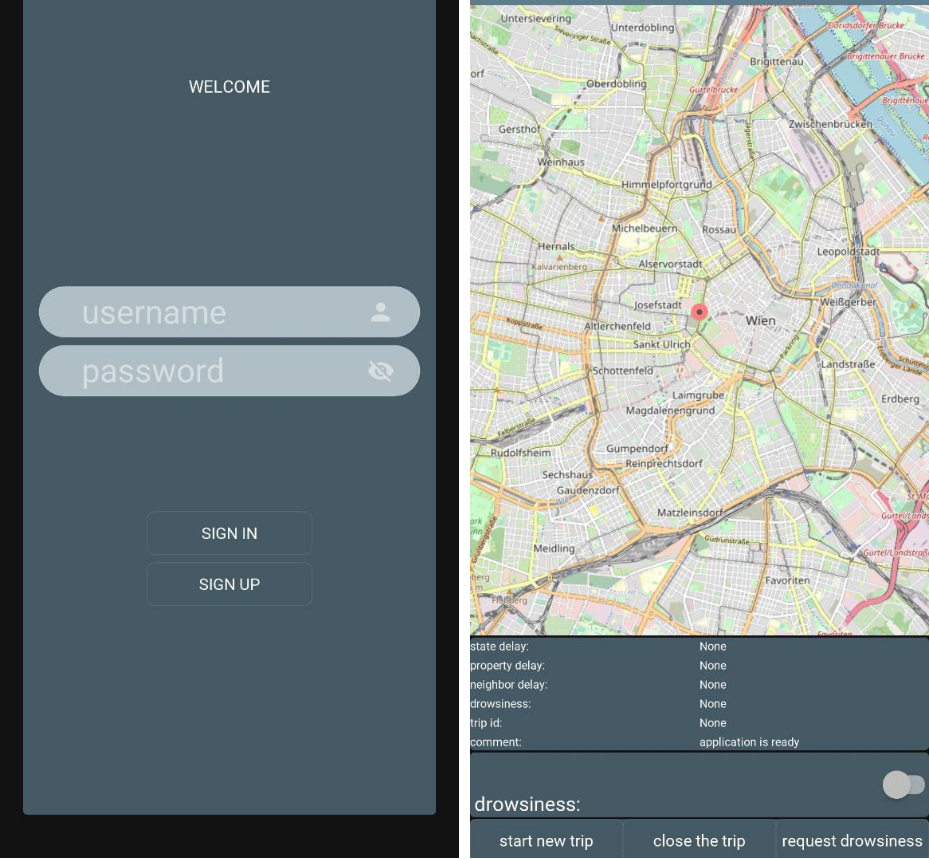}
	\caption{Test client application for android cellphones used during the field test} 
	\label{fig:app}
\end{figure}

We used the ``kivy'' framework for android phones to develop the client application. The application consisted of two screens (Figure~\ref{fig:app}). The first was a log-in page, the second was implemented for interacting and visualizing data from the server. After logging in and starting a trip, the application simultaneously measured GPS coordinates, transmitted them to RUCS, and requested neighbors from the service. Neighbors were then displayed as red markers on the user's map. The drowsiness of a driver in a neighboring vehicle could be requested simply by pressing the ``request drowsiness'' button. \\
We additionally built an application for the JKU-ITS vehicle by relying on The Robot Operating System (ROS). The ROS node was implemented in python to  allow reusability of the code written for the android client application. \\
Each request to the server was logged into the applications' local memory to evaluate the results of the experiment.\\
During the experiment, a connection to the server was provided via 4G cellular networks. 4527 messages were transferred to get a list of neighbors, 19776 messages were sent to indicate the vehicles states and 93 ``drowsiness'' parameter requests were made. The time between request and response was measured for each message, being all data logged in the system. 

\section{Results}
\label{sec:Results}

\begin{figure}[!t]
	\centering
	\includegraphics[width=0.4\textwidth]{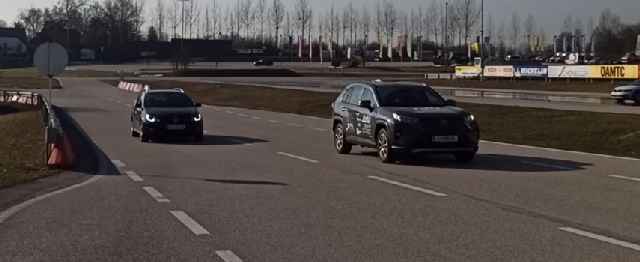}
	\caption{Screenshot of the two vehicles during the experiment} 
	\label{fig:jku-its-vehicle}
\end{figure}

\begin{figure}[!t]
	\centering
	\includegraphics[width=0.4\textwidth]{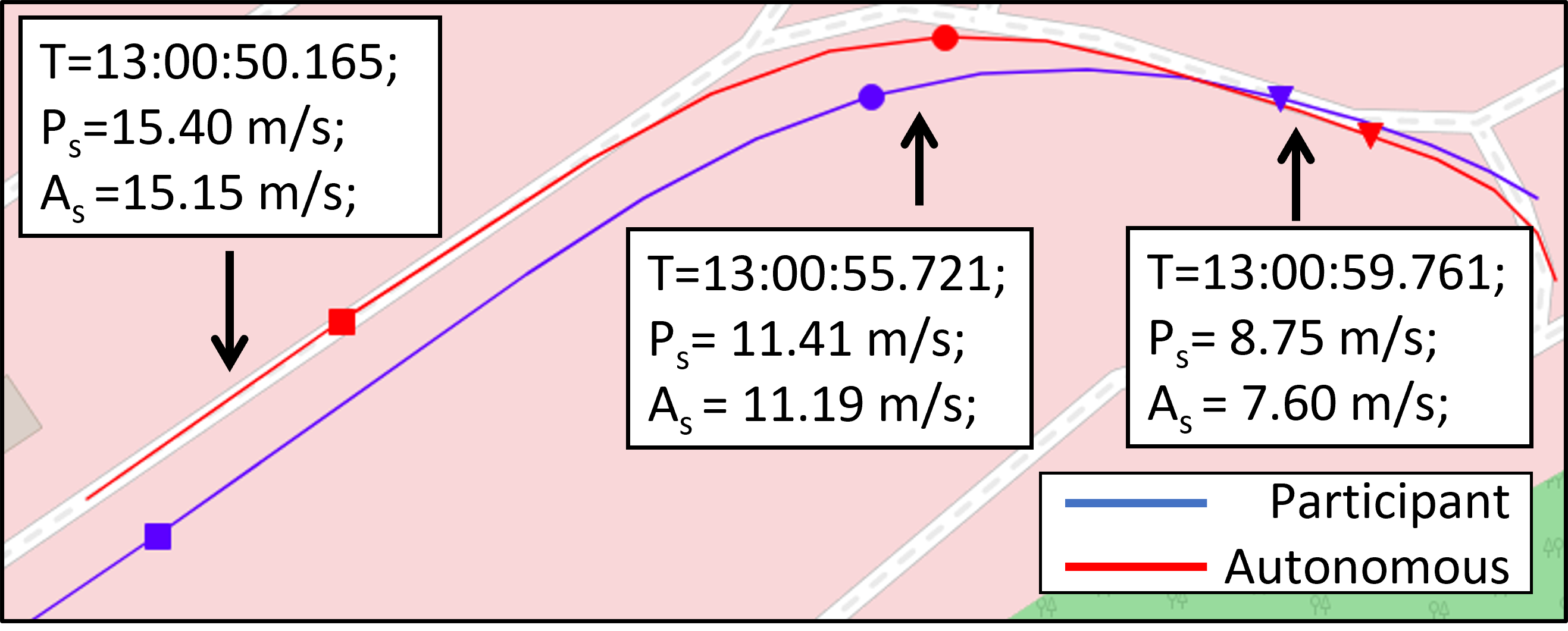}
	\caption{GPS data from vehicles during the experiment. The square, round and triangular signs represent the position of the vehicle A (in red) and B (in blue) before, during and after the drowsiness driver's state request respectively. As and Ps show the speeds for the vehicle A and vehicle B, respectively} 
	\label{fig:gps}
\end{figure}

\begin{figure}[!t]
	\centering
	\includegraphics[width=0.38\textwidth]{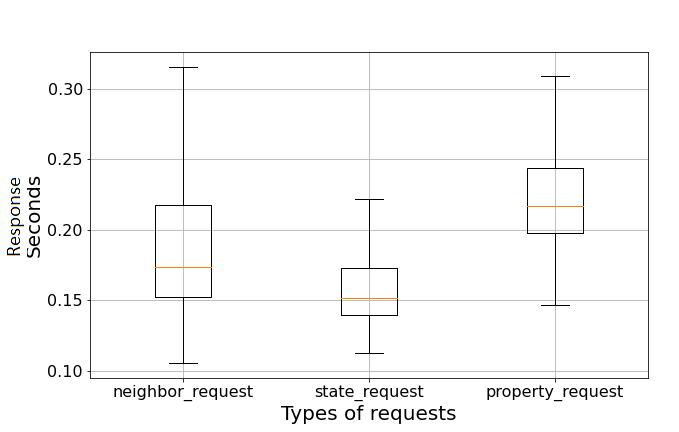}
	\caption{Distribution of the delays between the request and responds actions from the service} 
	\label{fig:time}
\end{figure}

\begin{figure}[!t]
	\centering
	\includegraphics[width=0.38\textwidth]{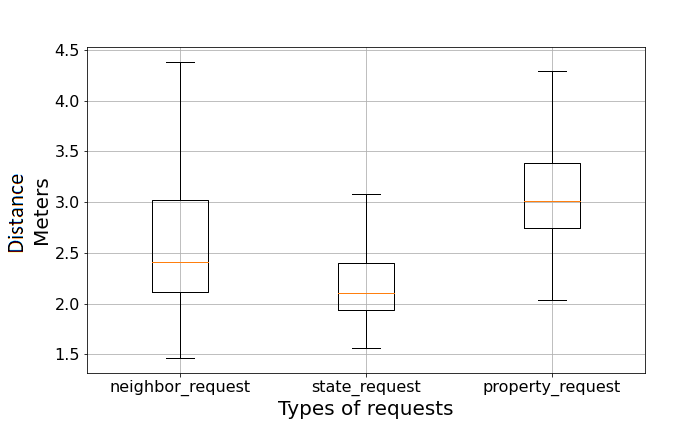}
	\caption{Distances traveled at a speed of 50 km/h. between the time a request is sent and the response is received from the service} 
	\label{fig:distance}
\end{figure}

The driven path for both vehicles is presented in Figure~\ref{fig:gps}. Initially, the autonomous vehicle A was slightly ahead of the manually-driven vehicle B (square signs). After a few seconds, the autonomous vehicle requested the drowsiness state from the vehicle B (round signs), and the server returned the correspondent value, in this case, low-level drowsiness. Finally, the vehicle A changed the lane in front of the vehicle B.
The average time for a response from the service did not exceed 0.25 seconds during the tests (Figure~\ref{fig:time}). Because the vehicles had an approximate speed of 50 km/h, a delay of 0.25 seconds equates to less than 3.5 meters of distance travelled between request and response from the server (Figure~\ref{fig:distance}). A delay of 0.25 seconds was quick enough for the service to be used comfortably as a communication platform in the use case investigated. 
Since the difference in the distribution of delays between the three requests was small, it can be concluded that most of the delay was due to the inherent delays in 4G cellular communication. Thus, it is reasonable to assume that the response delay will decrease in an environment with dense cellular coverage. 

\section{Conclusion and Future Work}
\label{sec:ConclusionAndFutureWork}

The coexistence of self-driving and manually-controlled vehicles is challenging and could negatively impact road safety. To facilitate their interaction, we presented in this paper a SaaS platform to enable communication between non-automated vehicles and vehicles with any level of automation. 

The validation results showed that the time delays of the service are low enough to not affect road situations in which time is not a critical safety factor. With the increasing improvement of cellular communication through 5G, these delays will be reduced. Delays can be further reduced by using a micro-service architecture in the application, which would enable the service to handle large amounts of data. Another improvement could be done by performing major calculations in advance based on the state transfer event and storing the results in a cache. Future work will target these improvements and the validation of the performance of the system under high load information.



\section*{ACKNOWLEDGMENT}

This work was supported by the Austrian Science Fund (FWF), within the project ''Interaction of autonomous and manually-controlled vehicles (IAMCV)'', number P 34485-N.

\bibliographystyle{IEEEtran}
\bibliography{lib}

\end{document}